\documentclass[a4paper,fleqn]{cas-sc}

\usepackage[authoryear]{natbib}

\usepackage{multirow}
\usepackage{colortbl}
\usepackage{booktabs}
\usepackage{pifont}

\def\tsc#1{\csdef{#1}{\textsc{\lowercase{#1}}\xspace}}
\tsc{WGM}
\tsc{QE}

\begin{document}
\let\WriteBookmarks\relax
\def\floatpagepagefraction{1}
\def\textpagefraction{.001}

\shorttitle{}    

\shortauthors{}

\title [mode = title]{Temporal2Seq: A Unified Framework for Temporal Video Understanding Tasks}

\author[1]{Min Yang}[orcid=0009-0007-8298-8161]



\ead{yangminmcg1011@hotmail.com}


\credit{Conceptualization, Methodology, Writing - original draft}

\affiliation[1]{organization={State Key Laboratory for Novel Software Technology, Nanjing University},
            city={Nanjing},
            postcode={210023}, 
            country={China}}

\author[1]{Zichen Zhang}


\ead{zichenzhang@smail.nju.edu.cn}


\credit{Investigation, Methodology,
Writing - review \& editing}

\author[2]{Qian Dang}


\ead{2937222047@qq.com}


\credit{Funding acquisition, Supervision, Validation,
Writing - review \& editing}

\author[1]{Limin Wang}[orcid=0000-0002-3674-7718]

\cormark[1]


\ead{lmwang@nju.edu.cn}


\credit{Supervision, Writing - review \& editing}

\affiliation[2]{organization={China Design Group Co., Ltd},
            city={Nanjing},
            postcode={210014}, 
            country={China}}

\cortext[1]{Corresponding author}


\begin{abstract}
The rapid advancement of video understanding has led to a proliferation of clip-level temporal analysis tasks,
 including temporal action detection (TAD), temporal action segmentation (TAS), and generic event boundary detection (GEBD). While task-specific video understanding models have exhibited outstanding performance in each task, there remains a lack of a unified framework capable of simultaneously addressing multiple tasks, which is a promising direction for the next generation of AI. To this end, in this paper, we propose a single unified unimodal framework, coined as \textbf{Temporal2Seq}, to formulate the output of these temporal video understanding tasks as a sequence of discrete tokens. With this unified token representation, Temporal2Seq can train a generalist model within a single architecture on different video understanding tasks. In the absence of multi-task learning (MTL) benchmarks, we compile a comprehensive co-training dataset by borrowing the datasets from TAD, TAS, and GEBD tasks. Our Temporal2Seq can produce reasonable results on various tasks and achieve advantages compared with the baseline of each task. It also shows better generalization performance on new datasets from different tasks, which yields superior performance to the specific model.
\end{abstract}




\maketitle

\section{Introduction}\label{intro}

In recent years, the video understanding community has witnessed a proliferation of video understanding tasks and their associated datasets in different scenarios such as temporal action detection (TAD)~\citep{thumos14,fineaction}, temporal action segmentation (TAS)~\citep{breakfast,gtea,50salads}, and generic event boundary detection (GEBD)~\citep{kinetics-gebd,tapos}. Meanwhile, many task-specific models~\citep{actionformer,diffact,temporalperceiver,ddm,asformer,react,tadtr} have achieved astonishing results in their tasks. With the emergence of large language models (LLMs)~\citep{chatgpt}, more and more different language-related tasks are beginning to be unified into a modeling framework. Inspired by this, temporal video understanding community also demands the unification of multiple tasks with the rapid growth of temporal video data~\citep{ego4d,epic-kitchen}. However, these task-specific models~\citep{actionformer,diffact,temporalperceiver} above cannot handle different temporal understanding tasks.

\begin{figure*}
    \centering
    \includegraphics[width=\textwidth]{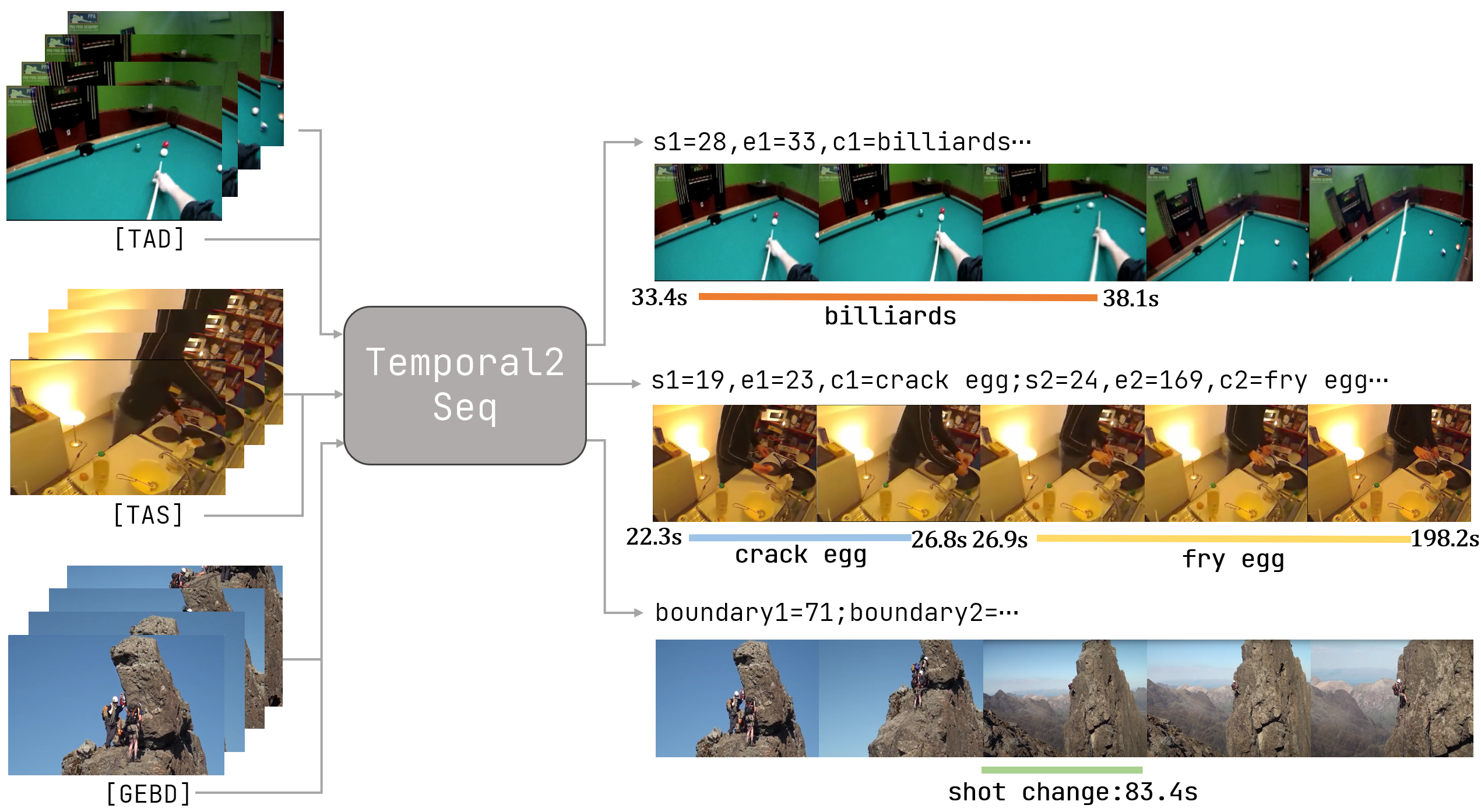}
    \caption{\textbf{The overview of Temporal2Seq.} We input video sequences from different tasks and their corresponding task prompts $[TASK]$ into the model, the model produces task output tokens which can be detokenized into the required task output for visualization.}
    \label{fig:temp2seq}
\end{figure*}

Multi-Task Learning (MTL)~\citep{yolor,pix2seq_v2,uni_perceiver_v2} has consistently been one of the most popular techniques to solve such challenges. It aims to utilize a single model to train on multiple tasks, jointly improving all tasks across various fields. For example, large language models~\citep{chatgpt} have successfully trained a single unified model to handle all downstream tasks via an autoregressive architecture. Meanwhile, some works such as Pixel2Seq V2~\citep{pix2seq_v2} or Unified IO~\citep{unified_io} have verified the effectiveness of multi-task learning on image understanding tasks, which further inspired related research in the video field~\citep{unloc,univtg,vid2seq,unimd}. However, they attempted to use the text modality to unify coarse-grained tasks like temporal video grounding and dense video captioning. 
We define 'coarse-grained' tasks as those involving long-duration actions (sometimes lasting several minutes) that occur sparsely within a video, making them relatively easier to detect. Moreover, they failed to address fine-grained tasks with associated datasets that do not rely on the text modality and require a deeper understanding of temporal visual context which needs more refined predictions. For example, TAS requires finely segmenting the entire video, GEBD demands accurately identifying each generic boundary, and TAD involves detecting all potential actions throughout the video instead of only one action corresponding to the text query. Therefore, no work has yet attempted to incorporate the above three tasks into constructing a unified model.

To this end, we build a single unified unimodal framework, termed as {\bf Temporal2Seq}, for different kinds of temporal video understanding tasks with a sequence-to-sequence architecture. As a proof of concept, we choose three tasks, including temporal action detection (TAD), temporal action segmentation (TAS), and generic event boundary detection (GEBD). Inspired by the Pix2Seq V2~\citep{pix2seq_v2}, our Temporal2Seq framework formulates the output of these three video understanding tasks as a sequence of discrete tokens, as shown in Figure ~\ref{fig:temp2seq}. This unified token representation endows our Temporal2Seq with a simple and general interface to handle three tasks within a single framework jointly. To benchmark the performance of Temporal2Seq, we compile a comprehensive benchmark of temporal action understanding tasks by borrowing the datasets from each task and co-train our Temporal2Seq model on these datasets.
After training, our single generalist model can perform different video understanding tasks via a simple prompt ([TASK]). The experiment results demonstrate that our single Temporal2Seq model outperforms the baseline counterparts for three tasks. To further investigate the advantage of our Temporal2Seq generalist model, we transfer this unified model to new datasets from different tasks to test its generalization ability, yielding superior performance to the specific model. In summary, our contributions are as follows:
\begin{itemize}
    \item We propose a single, unified framework for handling temporal video understanding tasks. To our knowledge, our Temporal2Seq is the first unified video modeling framework for handling different types of fine-grained temporal video understanding tasks without text modality.
    \item We successfully co-train our Temporal2Seq model on three diverse video understanding tasks, covering detection, segmentation, and timestamp localization. 
          Temporal2Seq can be flexibly applied to different tasks via a simple task prompt.
    \item Our Temporal2Seq empirically demonstrates the improvement of co-training across all three tasks over each specific model. Our generalist model also achieves competitive performance to established task-specific models under fair conditions. We also show its promising generalization ability on new datasets from these tasks.
\end{itemize}

\section{Related Work}
\noindent\textbf{Temporal Action Detection.} Temporal action detection (TAD) aims to localize the temporal interval of each action instance in an untrimmed video and recognize its action category. Existing TAD methods can be divided into one-stage TAD and multiple-stage TAD. One-stage TAD methods~\citep{actionformer,tadtr,basictad,pbrnet,daotad,react} aim to detect the boundaries and categories of action segments in a single shot while multi-stage TAD methods~\citep{bsn,bmn,bsn++,rtd,vsgn,gtad,afsd} often involve multiple stages to generate and refine action detection results. These aforementioned works attempt to solve the TAD task's localization and regression problems using different architectures, and tend to split localization and regression into two parallel or progressive branches. However, they have not attempted to use the same module to address both action regression and classification problems. 
Our Temporal2Seq converts action boundaries and class labels into sequences of discrete tokens to generate action predictions. Its unified architecture makes it possible to combine the TAD task with other temporal video understanding tasks.

\noindent\textbf{Temporal Action Segmentation.} Temporal action segmentation (TAS) aims to classify actions in untrimmed videos and provide frame-by-frame action label predictions. Some early work~\citep{DBLP:conf/cvpr/RohrbachAAS12, karaman2014fast} migrated methods from TAD to TAS tasks through sliding windows and non-maximum suppression. Other works use Markov models~\citep{kuehne2016end, tang2012learning} or RNNs~\citep{donahue2015long, yeung2018every} to model the temporal sequence and then classify framewise actions. 
With the rise of temporal convolutional networks~\citep{tcn} and Transformer~\citep{attention}, numerous outstanding works have emerged~\citep{mstcn,mstcn++,asformer,uvast,diffact}. They adopt multi-stage improvement modules to refine predictions and achieve significant success. Our Temporal2Seq adopts a dense prediction paradigm to output the action category of each frame in the form of discrete tokens. These frame-level predictions are then transformed into segment-level action segmentations without any post-processing algorithms. Although we do not use multi-stage improvement modules in TAS~\citep{uvast,rtk,fact}, Temporal2Seq can still achieve reasonable segmentation results.

\noindent\textbf{Generic Event Boundary Detection.}
Generic event boundary detection (GEBD) aims to locate the general boundaries that divide videos into semantically coherent and taxonomy-free units and could serve as an important pre-processing step for clip-level video understanding. Previous GEBD methods~\citep{pc,tcn,bsn,bmn,ddm} focused on building representations specifically designed for event-level boundaries and exploited a dense prediction paradigm with postprocessing. Temporal Perceiver~\citep{temporalperceiver} tried a sparse prediction paradigm by constructing boundary queries that directly regress the location of event boundaries. 
Our Temporal2Seq follows the dense prediction paradigm by predicting whether the current frame is an event boundary.

\noindent\textbf{Multi-Task Learning.}
The goal of Multi-task learning (MTL) is to train a single model to learn multiple tasks simultaneously. Such approaches offer several advantages including improving data efficiency, reducing overfitting through shared representations, and fast learning by leveraging auxiliary information. Existing unified video understanding models~\citep{unloc,univtg,vid2seq,unimd} introduced text modality and unified temporal video grounding, dense video captioning, and other multi-modal tasks. However, they failed to solve tasks that do not rely on the text modality and require a deeper understanding of temporal visual context. Also, they only solve coarse-grained video understanding tasks which tend to align the video and text modalities, leveraging large models' understanding of text semantics to solve video understanding tasks. Without the introduction of the text modality, these models would not work. Currently, there is no multi-task visual-only learning pipeline focusing on TAD, TAS and GEBD. 
We follow Pix2Seq V2~\citep{pix2seq_v2} to build a multi-task autoregressive pipeline on three important video understanding tasks, using one modal to solve them all. Although our model does not use any complex improvement modules and is merely a baseline, it achieves better results than using a single model by leveraging datasets from different sources across multiple scenes.

\begin{figure*}
    \centering
    \includegraphics[width=\textwidth]{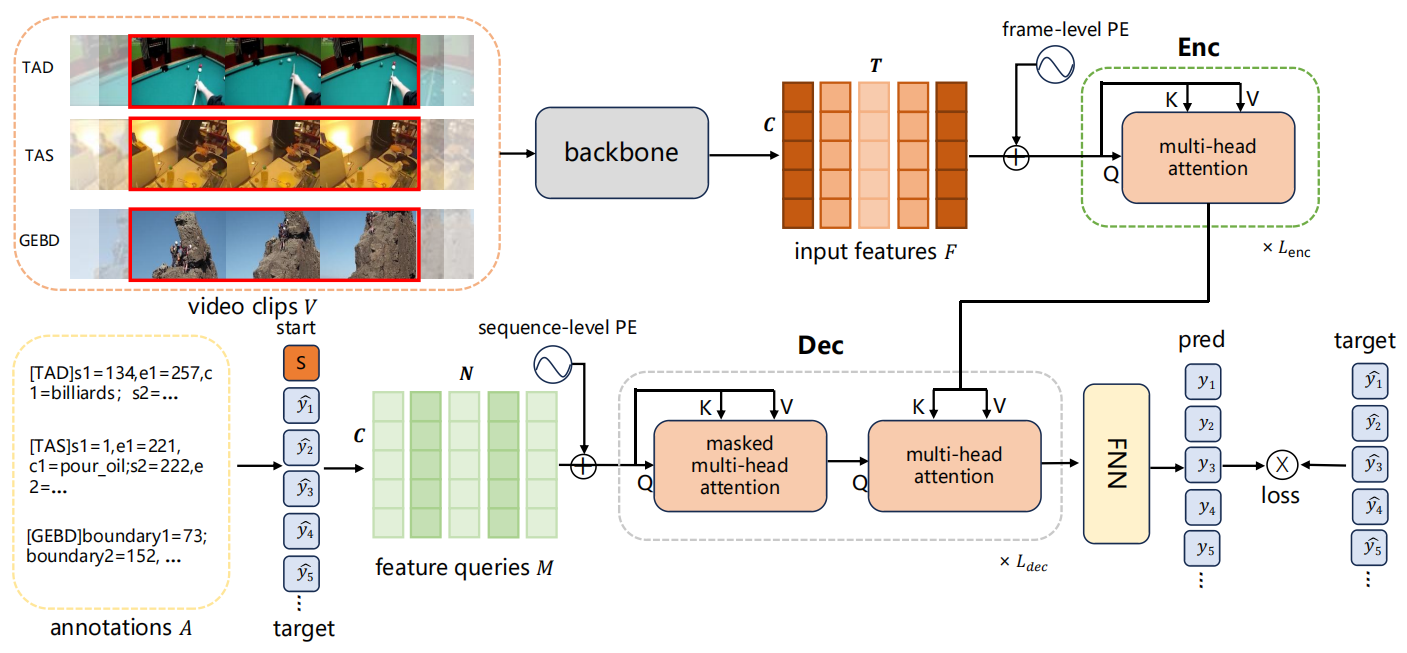}
    \caption{
        \textbf{The overall pipeline of Temporal2Seq.} The input to our model is video features with temporal dimension $T$ extracted by the backbone and a sequence of discrete tokens with token numbers of $N$ translated from annotations. Added with frame-level positional encoding, the encoder maps them into hidden representations. At training time, the decoder takes feature queries $M$ transformed from task annotations $A$ as input and predict the output conditioned by prompt start token, and a loss function is applied afterward. During inference, the decoder generates one token at a time conditioned on the preceding tokens and keeps repeating until the model provides all predictions. Due to space limitations, $H$ and $O$ are not shown here.
    }
    \label{fig:Temporal2Seq}
\end{figure*}

\section{Method}

\subsection{Overview}

The overall pipeline of Temporal2Seq is depicted in Figure~\ref{fig:Temporal2Seq}, which unifies these three different temporal video understanding tasks into a sequence-to-sequence framework. Given video clips $V$ sampled to the same length for joint training from untrimmed videos of three tasks along with annotations $A$ for each task. We encode $V$ into frame-level features from backbone~\citep{videomae_v2} and transform them into a latent feature space of reduced dimension $F \in R^{T \times C}$ as input features, where $T$ is the number of frames and $C$ is the feature dimension. Similar to Transformer~\citep{attention}, we add frame-level positional encoding to represent its temporal order. Then the Encoder $Enc$ consisting of $L_{enc}$ layers transforms the input features into hidden representations $H \in R^{T \times C}$ for Decoder $Dec$ consisting of $L_{dec}$ layers. As to annotations $A$, We introduce time tokens representing relative timestamps or boundaries and class tokens representing action categories. We formulate them into the target token sequence and add a start token $s$ ($[TAD]$, $[TAS]$ and $[GEBD]$) to the sequence and embed these discrete tokens into query embeddings $M \in R^{N \times C}$ via a dictionary look-up, where $N$ represents the number of target tokens. We add sequence-level positional encoding to represent the order of sequence and then send them into $Dec$ to generate output embeddings $O \in R^{N \times C}$. After that, a feed-forward network (FFN) maps $O$ back to predicted tokens. The model is trained to maximize the likelihood of token prediction conditioned on previous target tokens with a cross-entropy loss.

\begin{figure*}
    \centering
    \includegraphics[width=\textwidth]{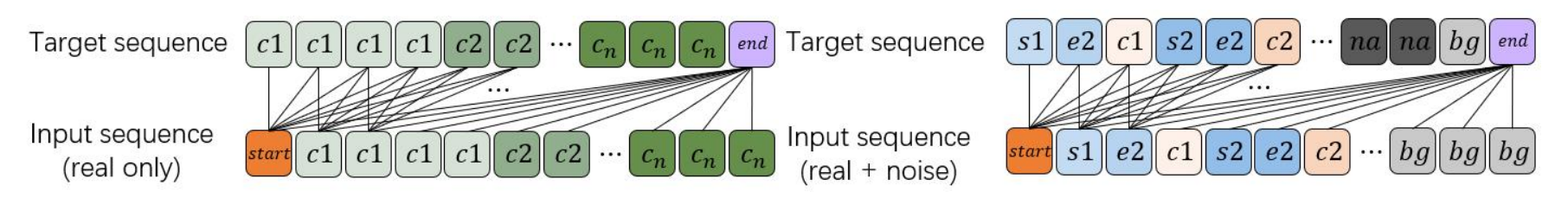}
    \caption{
        \textbf{Dense and sparse autoregressive paradigm.} The dense autoregressive paradigm aims to predict frame-by-frame categories ($c_{i}$), where $c_{i}$ can represent all the roles including the category and background. Following the settings in~\citep{pix2seq_v2}, input tokens in sparse autoregressive paradigm are constructed in triplet form representing the boundary ($s_{i}$ and $e_{i}$) and category ($c_{i}$) of the action. Since triplet targets cannot cover the whole prediction space, we fill them with "$n/a$" token and "bg" token during training. We also set the loss weight of "$n/a$" tokens to 0.1 to guide the model in not mimicking them. "bg" is short for background and it also represents the noise token for TAD. 
    }
    \label{fig:tad_tas_sequence}
\end{figure*}

\begin{figure*}
    \centering
    \includegraphics[width=\textwidth]{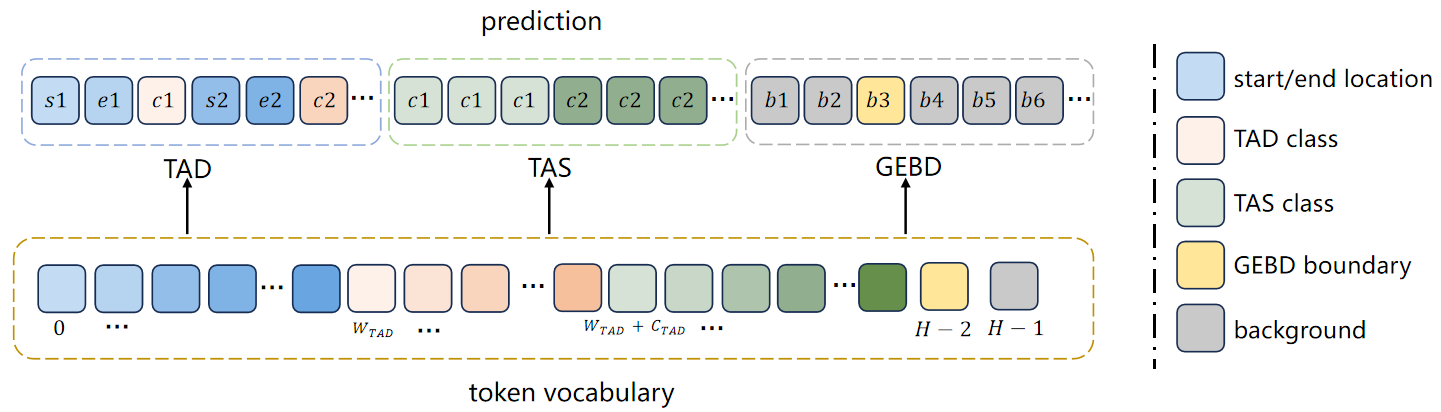}
    \caption{
        \textbf{Construction of token vocabulary.} We allocate location tokens for action boundaries and category tokens for all three tasks. During inference, the model generates output tokens, each token corresponds to a position in vocabulary.
    }
    \label{fig:sequence}
\end{figure*}

\begin{figure*}
    \centering
    \includegraphics[width=\textwidth]{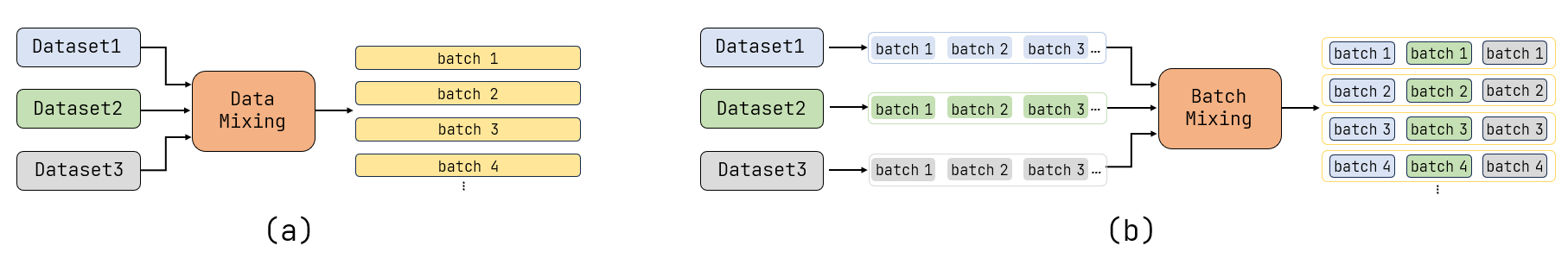}
    \caption{
        \textbf{Mixing ways of training datasets.} (a) Data mixing involves the creation of a dataset that contains mixed frame-target sequence pairs drawn from different tasks and then split into batches for each iteration. (b) Batch mixing samples batches of data from all tasks and then trains the combined batches in each iteration. 
    }
    \label{fig:mixing}
\end{figure*}

\subsection{Unified Interface with Tokenization}

While TAD, TAS, and GEBD are tasks related to video understanding, they are diverse and traditionally formulated quite differently. TAD needs to localize the start and end of each action and classify their categories. TAS requires the model to generate a dense frame-wise mask for each identified action instance. GEBD has to localize generic event boundaries rather than segment-level prediction. To solve these tasks using a single model rather than designing separate detectors for each task as in other works~\citep{univtg,unloc}, we should provide a unified interface to the task inputs and outputs. We design such vocabulary with the length of $H$ and employ a consistent color system to represent the type of token shown in Figure~\ref{fig:sequence}.
We put time tokens representing TAD's action boundaries, TAD's action categories, TAS's action categories, GEBD boundary and background into the token vocabulary. Assuming that we normalize time tokens into [0,$W_{TAD}$) as boundary interval, which means that the action boundary has $W_{TAD}$ values. Next, the token vocabulary stores the action categories of TAD and TAS. The token value range of these categories is [$W_{TAD}$,$W_{TAD}+C_{TAD}+C_{TAS}$) where $C_{[TASK]}$ represents the total number of categories in one task. We place the GEBD boundary and background classes in the last two positions of the token vocabulary for simplicity in code. The length of $H$ should be greater than or equal to the number of tokens required to be encoded. 
Specifically, a prediction for TAD can be represented as $(s_i,e_i,c_i)$ which represents the start boundary $s$, the end boundary $e$ and action's category $c$ for the $i^{th}$ action, we give consecutive triples like above as predictions as shown in Figure~\ref{fig:tad_tas_sequence}.
For TAS, we follow previous works~\citep{asformer,ddm} and build a dense prediction paradigm without post-processing algorithms. Formally, we classify action categories frame by frame and get predictions $\{c_i\}|^{L}_{i=1}$ where $L$ represents the length of the video frames, and then stitch these per-frame predictions into segment-level action segmentation predictions.
For GEBD, it turns into a per-frame binary classification problem, determining whether it is an action boundary or a background frame by frame and getting predictions as $\{b_i\}|^{L}_{i=1}$ where $b_i$ represents whether there is a boundary located at $i^{th}$ temporal location. Then we convert it into a boundary prediction result $\{boundary_i\}|^{N}_{i=1}$ where $N$ is the number of detected boundaries as shown in Figure~\ref{fig:tad_tas_sequence}. In summary, we adopt dense autoregressive paradigm for TAS and GEBD and sparse autoregressive paradigm for TAD.

\subsection{Training}

In this section, we will discuss how to train Temporal2Seq on three tasks jointly. We first describe our training strategies, and then introduce our loss function design for each task, especially a new loss function for TAD. 

\noindent\textbf{Two Ways of Joint Training.}
Inspired by Pix2Seq V2~\citep{pix2seq_v2}, we adopt the following two ways for co-training on different tasks and datasets shown in Figure~\ref{fig:mixing}.
In data mixing settings, the datasets of all tasks will be mixed and divided into multiple groups according to batch size as shown in Figure~\ref{fig:mixing}(a). All these groups will be trained only once within an epoch and a batch of data from a sampled group will be trained in each iteration. After that, we compute the loss and update this model. We use \textbf{Temporal2Seq$_{data}$} to represent the model trained in this way. In batch mixing settings, we pre-partition the dataset into groups based on batch size configuration for each task as shown in Figure~\ref{fig:mixing}(b). After that, a fixed number of batches are selected from these groups and then spliced together for training. It is worth noting that datasets with fewer groups will be input cyclically within the epoch until the datasets with more groups finish their training process. Therefore, the final batch IDs of these datasets will be different shown in Figure~\ref{fig:mixing}(b). We use \textbf{Temporal2Seq$_{batch}$} to represent the model trained in this way.

\noindent\textbf{Data Balance Strategy.}
In the training process, we observe discrepancies stemming from imbalanced dataset scales and substantial variations in the difficulty levels associated with training tasks across different datasets. Specifically, GEBD's task difficulty is lower than that of TAS and TAD, and its excessive data can easily disrupt the training of the other two. This means that only the GEBD task can benefit from joint training, while the results of other tasks will be significantly degraded. Here, we propose the data balance strategy by discarding part of the data from the GEBD dataset in advance during training. 
Balancing datasets from different tasks allows joint training to learn a more general visual representation without bias towards a certain task. We first try this manner on \textbf{Temporal2Seq$_{batch}$} and stop training when the TAS and TAD datasets are fully trained in one epoch. We then apply this idea to \textbf{Temporal2Seq$_{data}$} and align it with the batch-mixing configuration by directly sampling GEBD data before training. During training, each epoch samples randomly from the dataset to enhance the robustness of data utilization. This training strategy can also be applied when introducing datasets from other tasks. If the imbalance in dataset sizes significantly affects other tasks, this data balancing strategy can be applied by truncating the imbalanced data during training and stopping the training on that dataset.

\noindent\textbf{Loss Functions.}
Although we use the same architecture for joint training, we apply different loss functions for each task. In the original Pix2Seq V2~\citep{pix2seq_v2}, the unified task was transformed into a standard token classification task. However, this approach can be brutal for TAD, as any prediction that does not regress to the correct location is penalized equally. To relieve this issue, we impose more penalties on predictions further from the ground truth boundaries by building a simple loss function called \textbf{weight loss} $L_{w}$ for TAD: 
\begin{equation}
        \begin{aligned}
            L_{w}(t) =
            \begin{cases}
                -\log(y_{t,\hat{c}}) \,\,\quad\quad\qquad\qquad\qquad\qquad\quad t \in \mathcal{T}_{cls} \\
                -(1+\frac{|argmax(y_{t,\hat{c}})-\hat{c})|}{D})\log(y_{t,\hat{c}}) \,\quad\qquad t \in \mathcal{T}_{bnd}
            \end{cases}
        \end{aligned}
    \label{eq:example}
\end{equation}

where $y_{t,\hat{c}}$ is the predicted probability for the ground truth label $\hat{c}$ at sequence token $t$. $\mathcal{T}_{cls}$ and $\mathcal{T}_{bnd}$ clearly denote the subset of indices corresponding to the class predictions and boundary predictions, respectively.
When output action category prediction, $L_{w}$ is the same as cross-entropy loss. When output action's boundary prediction, we need to impose a corresponding penalty based on the distance between the predicted position and the ground truth. We use $argmax(y_{t,\hat{c}})$ to calculate the predicted boundary while the ground truth is $\hat{c}$. $D$ represents the length of boundary space.

For TAD, we adopt weight loss as: 

\begin{equation}
    \begin{aligned}
        L_{tad} = \frac{1}{T}\sum_{t}L_{w}(t), \label{eq:l_tad}
    \end{aligned}
\end{equation}

For GEBD, we use the classification loss $L_{gebd}$ as:
\begin{equation}
    \begin{aligned}
        L_{gebd} = \frac{1}{T}\sum_{t}-\log(y_{t,\hat{c}}),
    \end{aligned}
\end{equation}
where $y_{t,\hat{c}}$ is the predicted probability for ground truth $\hat{c}$, $T$ represents the number of input frames, $t$ represents the current frame.

For TAS, we further adopt smooth loss which is widely used in TAS methods~\citep{asformer} to prevent over-segmentation. the loss function $ L_{tas}$ is a combination of classification loss $L_{cls}$ for each frame and smooth loss $L_{smo}$~\citep{mstcn} which calculates the mean squared error over the frame-wise probabilities:
\begin{equation}
    \begin{aligned}
         & L_{tas} = L_{cls} + \lambda L_{smo} & = \frac{1}{T}\sum_{t}-\log(y_{t,\hat{c}})
        + \lambda \frac{1}{TC}\sum_{t}\sum_{c}(y_{t-1,c}-y_{t,c})^{2}, \label{eq:l_tas}
    \end{aligned}
\end{equation}
where $y_{t,\hat{c}}$ is the predicted probability for ground truth $\hat{c}$ at frame $t$, $y_{t,c}$ is the predicted probability for label $c$ at frame $t$ and $C$ is the total number of action categories. $\lambda$ represents the weight for $L_{smo}$.

With these loss functions, we co-train our Temporal2Seq models \textbf{Temporal2Seq$_{batch}$} in batch-mixing setting and \textbf{Temporal2Seq$_{data}$} in data-mixing setting. We also use the above loss to train the individual model \textbf{Baseline$_{[TASK]}$} corresponding to each task.

\subsection{Inference}
During inference, our Temporal2Seq takes the video frames and the corresponding task prompt as input shown in Figure~\ref{fig:sequence}. 
Our Temporal2Seq starts from the corresponding task prompt token and generates predictions in a sequence format from the model likelihood, i.e., $P(y_j|F,y_{1:j-1})$. 
For TAD, Temporal2Seq outputs detection predictions as a list of triplets. We start with $[TAD]$ as the start token and give triplet predictions based on input video clips, formulated as $(s_i, e_i, c_i)$. Since our model outputs a fixed number of triplets, it inevitably produces redundant predictions. Therefore, we apply Non-Maximum Suppression (NMS) to filter them.
For TAS, Temporal2Seq outputs each frame's action category and combines them into segments. We start with $[TAS]$ as the start token and give frame-by-frame action category predictions. They are then converted into dense segment-level segmentation results in the form of triplets, where $e_i = s_{i+1}-1$.
For GEBD, Temporal2Seq outputs each frame's binary predictions of generic boundaries, and then convert them into locations of boundaries. We start with $[GEBD]$ as the start token and output each frame’s binary predictions. The token position predicted as a boundary will be converted into an action boundary. Specifically, we give the location prediction of the boundary based on its relative position within the temporal sampled window.

\section{Experiments}
\subsection{Datasets and Evaluation Metrics}
We select \textbf{THUMOS14}~\citep{thumos14} and \textbf{FineAction}~\citep{fineaction} as our choices for TAD. For evaluation, we report the mean average precision (mAP) at different temporal intersections over union (tIoU) thresholds [0.3:0.1:0.7] for THUMOS14 and [0.5:0.05:0.95] for FineAction. Avg is the average mAP on these thresholds.
We adopt two widely used TAS datasets~\citep{breakfast,gtea} called \textbf{Breakfast}~\citep{breakfast} and \textbf{GTEA}~\citep{gtea}. We use frame-wise accuracy (Acc), segmental edit score (Edit), and segmental overlap F1 score with threshold k/100, denoted as F1@k, to evaluate the performance.
In order to align with other tasks, we do \textbf{NOT} perform 5-fold cross-validation adopted by other TAS works~\citep{asformer,mstcn,diffact}, and only report results from the first split.
For GEBD, we adopt \textbf{Kinetics-GEBD}~\citep{kinetics-gebd} and \textbf{TAPOS}~\citep{tapos}. We use the F1 score under different Relative Distance thresholds [0.05 : 0.05 : 0.5] for quality measurement. Avg is the average F1 scores on these thresholds.  Due to space limitation, we only report F1 score with a threshold of 0.05 and their average F1 score in some tables. We adopt \textbf{THUMOS14}, \textbf{Breakfast}, and \textbf{Kinetics-GEBD} to co-train our Temporal2Seq because they are the most representative in each task, and other datasets will appear in our ablation studies. 

\subsection{Implementation Details}
For all datasets from each task, we adopt ViT-B from~\citep{videomae_v2} for feature extraction and the sampling stride $\tau = 1$ for THUMOS14, $\tau = 2$ Breakfast and $\tau = 1$ for Kinetics-GEBD. We randomly crop window clips of the same length to each video in all three datasets for batch training. If the size of the sampling window is larger than the actual length of the visual feature, pad the visual feature to the size of the sampling window. Sliding windows are used during inference to generate predictions.
In all experiments, we train the model for 1200 epochs with each epoch involving a randomly sampled clip from each video. The boundary space $D$ is 1200, combined with the fps and sampling stride $\tau$ of the respective datasets, we crop each video clip with temporal windows of 40 seconds, 160 seconds, and 10 seconds for THUMOS14, Breakfast, and Kinetics-GEBD.
During separate training of individual task model \textbf{Baseline$_{[TASK]}$}, we set batch size 4 for THUMOS14, 16 for Breakfast, and 16 for Kinetics-GEBD. When training them jointly, we set 4 batches for THUMOS14, 12 batches for Breakfast, 16 batches for Kinetics-GEBD for batch-mixing manner called \textbf{Temporal2Seq$_{batch}$}. In data-mixing settings, \textbf{Temporal2Seq$_{data}$} uses a batch size of 32.
We use AdamW~\citep{adamw} as optimizer and a learning rate of 2e-4 following settings from TP~\citep{temporalperceiver}. The weight $\lambda$ is set to 0.15 in Equation \ref{eq:l_tas}. $L_{enc} = L_{dec} = 3$. Video frames $L = 1200$. We train our model based on 4 Nvidia RTX 3090 GPUs for one day to reach 1200 epochs. 

\subsection{Ablation Study}
\noindent\textbf{Effectiveness of Weight Loss for TAD.}
We compare our \textbf{weight loss} with cross-entropy loss based on \textbf{Baseline$_{TAD}$} trained from THUMOS14, an individual task model for TAD. Shown in Table~\ref{table:weight_loss}. The use of weight loss has resulted in improved accuracy in detection results (from 61.3 to 62.3). 
This suggests that using simple classification loss to supervise detection tasks is not the optimal choice, the introduction of weight loss makes the boundaries more precise. Although such a sequence-to-sequence autoregressive framework can transform all perception tasks into a classification task in token space, it is necessary to recognize the distinctions between classification and regression tasks and design different supervision schemes. We hope that our improvement approach for the loss function can inspire future work to explore a loss function that can effectively handle both classification and regression tasks within a unified architecture.

\noindent\textbf{Study on the Sequence Length and Stride for Different Tasks.} 
Unlike image-level unified models, video-level unified models need to consider the varying lengths of different videos. Due to the lack of a unified training dataset for joint training scenarios, the datasets from different tasks vary in scene and length. To address this issue, we use sliding windows of the same sequence length during joint training, along with different downsampling stride to control the input length, since the temporal length of the input video equals the product of the sequence length and the downsampling stride. A longer sequence length increases the effective length of the input video but also introduces additional computational overhead. Using a stride allows for increasing the effective length of the input video with the same sequence length, but at the cost of losing many details in video features. We conduct experiments on sequence length across various tasks to investigate the impact of temporal feature length on task performance and examine the impact of different downsampling rates on the results. As shown in Table~\ref{table:tad_sequence}, with the same window size, denser visual features lead to better results, and it reaches the best when Length$\times$Stride is $1200\times1$ which means the sliding window covers 40s, longer than 99.8\% of the actions in the video. When we further use a bigger stride to obtain longer caption of temporal features, the performance sharply decreases. 
As shown in Table~\ref{table:tas_sequence}, longer and denser visual features lead to better results for TAS. This is because TAS uses a dense prediction paradigm, where each frame only focuses on the action prediction of the current frame, which is far less difficult than understanding the entire action interval. We find that $1200 \times 2$ meets the best results. We also find that while longer sequence inputs enable the model to capture longer-term video content and improve its performance, blindly increasing the stride can lead to a drop in model performance. Therefore, we can conclude that longer sequence lengths help the model understand long-term videos, but a larger stride leads to excessive loss of temporal details, resulting in degraded model performance. Since the temporal length of GEBD is far below 1200, this issue does not exist. In summary, we select the 1200$\times$1 configuration for TAD tasks and the 1200$\times$2 configuration for TAS tasks.

\begin{table*}[t]
    \caption{\textbf{Effectiveness of Weight Loss}. Here we build the baseline and report mAP with all tIoU thresholds for TAD.
    }
    \label{table:weight_loss}
    \centering
    \tabcolsep=0.01\textwidth
    \begin{tabular}{c|c|ccccc|c}
        \hline
       Model & Loss          & mAP@0.3           & mAP@0.4           & mAP@0.5           & mAP@0.6           & mAP@0.7           & Average mAP           \\ \hline
     \multirow{2}{*}{Baseline$_{TAD}$}  & cross-entropy &   79.0 &   73.4        &    65.2     &    52.2  &  36.7  &    61.3     \\
      &  weight loss   & \textbf{80.6} & \textbf{74.2} & \textbf{66.3} & \textbf{53.1} & \textbf{37.3} & \textbf{62.3} \\
        \hline
    \end{tabular}
\end{table*}

\begin{table*}[t]  
    \caption{\textbf{Study on sequence length and stride for TAD.} We set \textbf{Length} to represent the input visual feature sequence and \textbf{Stride} to represent the downsampling rate. }
    \label{table:tad_sequence}
    \centering
    \tabcolsep=0.01\textwidth
    \begin{tabular}{c|c|ccccc|c}
        \hline
      Model &  Length $\times$ Stride & mAP@0.3  & mAP@0.4  & mAP@0.5  & mAP@0.6  & mAP@0.7  & Average mAP  \\
        \hline
     \multirow{9}{*}{Baseline$_{TAD}$} &   $150 \times 4$         & 70.2 & 62.7 & 52.9 & 41.2 & 27.9 & 51.0 \\ 
      &   $300 \times 2$         & 72.2 & 64.2 & 54.7 & 43.4 & 29.3 & 52.8 \\
    & $600 \times 1$         & 73.0 & 66.0 & 56.3 & 45.6 & 31.8 & 54.5 \\ 
        \cmidrule(r){2-8} 
     &   $300 \times 4$         & 76.3 & 69.2& 58.2&45.8 & 32.1& 56.3\\
     &   $600 \times 2$         & 77.7 & 71.1 & 62.7 & 49.9 & 34.5 & 59.2 \\
     &   $1200 \times 1$ & \textbf{80.6} & \textbf{74.2} & \textbf{66.3} & \textbf{53.1} & \textbf{37.3} & \textbf{62.3} \\
     \cmidrule(r){2-8}
        &$1200 \times 2$      &       77.3    & 71.8 & 63.3 & 50.5 & 35.7 &59.7 \\
        &$1200 \times 4$         &   75.1   & 68.2   & 58.5 & 46.1 & 32.8 & 56.1 \\
        &$1200 \times 8$        &    71.8   & 66.1   & 56.6 & 45.0 & 30.4 & 54.0 \\ 
        \hline
    \end{tabular}
\end{table*}

\begin{table*}[t]
    \caption{\textbf{Study on sequence length and stride for TAS.} We set \textbf{Length} to represent the input visual feature sequence and \textbf{Stride} to represent the downsampling rate.}
    \label{table:tas_sequence}
    \centering
    \begin{tabular}{c|c|ccccc}
        \hline
        Model & Length $\times$ Stride & F1@{10} & F1@{25} & F1@{50} & Segmental Edit Score & Frame-wise Accuracy                \\ \hline
        \multirow{9}{*}{Baseline$_{TAS}$} &$150 \times 4$         & 63.8                              & 57.3 & 47.2 & 63.1 & 58.3 \\ 
        &$300 \times 2$         & 64.9                              & 60.1 & 49.0 & 64.7 & 59.6 \\
        & $600 \times 1$         & 68.7                              & 62.9 & 51.1 & 65.5 & 60.6 \\ 
        \cmidrule(r){2-7} 
        &$300 \times 4$         &                   70.8        & 64.7 & 50.4 & 67.9& 63.2 \\
        &$600 \times 2$         &         72.7      & 66.8& 52.6 & 69.3& 64.8\\
        &$1200 \times 1$        &         73.9      & 68.0& 54.0 & 72.6& 65.4\\
        \cmidrule(r){2-7}
        &$1200 \times 2$      &           \textbf{75.7}        & \textbf{70.8} & \textbf{57.8} & \textbf{73.4} &  \textbf{70.7}\\ 
        &$1200 \times 4$         &                       74.7   & 70.3& 56.4 & 71.4& 69.5\\ 
        &$1200 \times 8$        &    73.2       & 69.2 & 55.1 & 70.3 & 67.8 \\
        \hline
    \end{tabular}
\end{table*}

\begin{table*}[t]
    \caption{\textbf{Study on prediction paradigm for TAD}. We apply both dense and sparse prediction paradigms on THUMOS14. 
    }
    \centering
    \tabcolsep=0.01\textwidth
    \begin{tabular}{c|c|c|cccccc}
        \hline
        Model & Length $\times$ Stride & Paradigm & mAP@0.3           & mAP@0.4           & mAP@0.5           & mAP@0.6           & mAP@0.7           & Avg.mAP           \\ \hline
    \multirow{2}{*}{Baseline$_{TAD}$} & \multirow{2}{*}{$1200\times1$} & sparse   & \textbf{80.6} & \textbf{74.2} & \textbf{66.3} & \textbf{53.1} & \textbf{37.3} & \textbf{62.3} \\
     &  & dense    & 67.6          & 58.2          & 48.0          & 35.6          & 21.4          & 46.2          \\ \hline
    \end{tabular}
    \label{table:dense_paradigm_tad}
\end{table*}

\begin{table*}[t]
    \caption{\textbf{Study on prediction paradigm for TAS}. Here we apply both dense and sparse prediction paradigms on Breakfast. 
    }
    \centering
    \tabcolsep=0.01\textwidth
    \begin{tabular}{c|c|c|ccccc}
        \hline
      Model & Length$\times$Stride & Paradigm & F1@10         & F1@25         & F1@50         & Segmental Edit Score         & Frame-wise Accuracy           \\ \hline
      \multirow{2}{*}{Baseline$_{TAS}$} & \multirow{2}{*}{$1200\times2$} & sparse   & 63.9          & 57.6          & 44.4          & 62.3          & 59.7          \\
     &  & dense    & \textbf{75.7} & \textbf{70.8} & \textbf{57.8} & \textbf{73.4} & \textbf{70.7} \\ \hline
    \end{tabular}
    \label{table:dense_paradigm_tas}
\end{table*}

\noindent\textbf{Study on Different Prediction Paradigms for Different Tasks.}
Here we attempt to apply the dense prediction paradigm for TAD. As shown in Table~\ref{table:dense_paradigm_tad}, the detection results are worse than the sparse paradigm by a large margin, indicating that the dense paradigm is unsuitable for temporal action detection. We also investigate the sparse paradigm for TAS. Since the segmentation task requires a unique prediction for each frame, when two predicted segments overlap, we resort to discarding the segment with the lower score in the overlapping region. As shown in Table~\ref{table:dense_paradigm_tas}, the segmentation results are also much worse than the dense paradigm by a large margin and coherent actions in TAS are not recognized. We find that each other's prediction paradigms are not applicable. Based on the above, we choose a sparse autoregressive paradigm for TAD and a dense autoregressive paradigm for TAS. 

\begin{table*}[tb]
    \caption{\textbf{Comparison with baseline models}. We produce baselines for each task to compare with our Temporal2Seq based on data mixing and batch mixing. We report mAP for THUMOS14, F1@\{10,25,50\}, Edit and Acc for Breakfast, F1 score with a threshold of 0.05 and their average F1 score for Kinetics-GEBD. 
    }
    \label{table:multi_vs_single}
    \centering
    \resizebox{1.0\textwidth}{!}{
        \begin{tabular}{ccccccccccccccc}
            \hline
            \multirow{2}{*}{Model} & \multirow{2}{*}{Param} & \multicolumn{6}{c}{TAD} & \multicolumn{5}{c}{TAS} & \multicolumn{2}{c}{GEBD} \\
            \cmidrule(r){3-8} \cmidrule(r){9-13}  \cmidrule(r){14-15} 
                                   &                        & 0.3                     & 0.4                     & 0.5                      & 0.6           & 0.7           & Avg           & \multicolumn{3}{c}{F1@\{10,25,50\}} & Edit          & Acc           & 0.05          & Avg                                           \\
            \hline
            Baseline$_{TAD}$       & 6.98M                  & \textbf{80.6}                    & 74.2                    & 66.3                     & 53.1          & 37.3          &62.3          & -                                   & -             & -             & -             & -             & -             & -             \\
            Baseline$_{TAS}$       & 6.98M                  & -                       & -                       & -                        & -             & -             & -             & 75.7                                & 70.8          & 57.8          & 73.4          & 70.7
                 
                                   & -                      & -                                                                                                                                                                                                                                                                  \\
            Baseline$_{GEBD}$      & 6.98M                  & -                       & -                       & -                        & -             & -             & -             & -                                   & -             & -             & -             & -             & 75.0          & 85.5          \\ \hline
          
            Temporal2Seq$_{batch}$ & 6.98M                  & 79.9           & \textbf{74.6}           & 67.4            & \textbf{56.4}          & \textbf{41.9} & \textbf{64.0} & \textbf{77.5}                       & \textbf{71.6} & 59.7 & \textbf{75.0}          &\textbf{72.2}          & 75.2          & 86.4 \\
        
            Temporal2Seq$_{data}$  & 6.98M                  &        80.3             & 74.2                   &   \textbf{67.7}                  & 55.2 &    40.3       & 63.5          &            77.0                    &    71.5       &   \textbf{60.3}        & 74.7 & 71.8 & \textbf{75.6} & \textbf{86.5} \\ \hline
        \end{tabular}
    }
\end{table*}

\begin{table*}[tb]
    \caption{\textbf{Study on data balance strategy for joint training}. Here we compare the results before and after data sampling of the GEBD dataset. Data sampling has a highly positive effect on the results of the other two tasks.}
    \label{table:data_sampling}
    \centering
    \resizebox{1.0\textwidth}{!}{
        \begin{tabular}{ccccccccccccccc}
            \hline
            \multirow{2}{*}{Model}                  & \multirow{2}{*}{Data Balance Strategy} & \multicolumn{6}{c}{TAD} & \multicolumn{5}{c}{TAS} & \multicolumn{2}{c}{GEBD}\\
            \cmidrule(r){3-8} \cmidrule(r){9-13}  \cmidrule(r){14-15} 
                                                    &                                        & 0.3                     & 0.4                     & 0.5                      & 0.6           & 0.7           & Avg           & \multicolumn{3}{c}{F1@\{10,25,50\}} & Edit           & Acc           & 0.05          & Avg                                           \\
            \hline
            \multirow{2}{*}{Temporal2Seq$_{batch}$} & \checkmark                             & \textbf{79.9}           & \textbf{74.6}           & \textbf{67.4}            & \textbf{56.4}          & \textbf{41.9} & \textbf{64.0} & \textbf{77.5}                       & \textbf{71.6} & \textbf{59.7} & \textbf{75.0}          &\textbf{72.2}          & \textbf{75.2}          & \textbf{86.4} \\
                                                    &                                        &72.3                    & 64.1                    & 54.2                     & 40.8          & 29.2          &    52.1       & 66.9                                & 56.1           & 48.7          & 64.2          & 60.4          & 74.8          & 85.7          \\ \hline
            \multirow{2}{*}{Temporal2Seq$_{data}$}  & \checkmark                             & \textbf{80.3}           & \textbf{74.2}           & \textbf{67.7}            & \textbf{55.2} & \textbf{40.3} & \textbf{63.5} & \textbf{77.0}                       & \textbf{71.5} & \textbf{60.3} & \textbf{74.7} & \textbf{71.8} & \textbf{75.6} & \textbf{86.5} \\
                                                    &                                        & 71.6                    & 63.4                    & 53.2                     & 40.2          & 28.7          &      51.4     & 66.6                                & 56.4           & 47.7          & 65.1          & 59.3          & 75.3          & 85.9          \\ \hline
        \end{tabular}
    }
\end{table*}

\noindent\textbf{Effectiveness of Co-training on Multiple Tasks.} 
In this section, we present the co-trained results of Temporal2Seq based on two ways of joint training and obtain \textbf{Temporal2Seq$_{data}$} and \textbf{Temporal2Seq$_{batch}$}. As shown in Table \ref{table:multi_vs_single}, all tasks achieve improved performance without introducing additional training parameters compared with \textbf{Baseline$_{[TASK]}$} for each task. We will analyze the reasons why joint training is effective from both the encoder and decoder perspectives. As to the encoder, it shares weights for all tasks. Different scenarios from each task enable the model to learn more generalizable contextual semantics. As to the decoder, equipped with sufficient mapping spaces for predictions, Temporal2Seq can be adopted to more datasets without being confused during training. Although our architecture does not rely on complex, task-specific modules, the shared encoder and decoder effectively leverage diverse datasets from various tasks and achieve better performance compared with individual models. 

\noindent{\bf{Study on Data Balance Strategy for Joint Training.}} 
As shown in Table \ref{table:data_sampling}, when we send all data from each task into our model during training, the results of both TAD and TAS are severely affected. This phenomenon prompts us to consider the varying levels of training difficulty and dataset sizes during joint training. To solve this issue, we attempt to maintain the data distribution from different tasks during joint training by discarding portions of data that were not fully trained after completing the training on data from other tasks during each epoch. This ensures that each task can benefit from shared data without being overly influenced by the data from other tasks. As shown in Table \ref{table:data_sampling}, this simple operation results in a significant improvement, as it can effectively balance the training for various tasks without adversely affecting GEBD's results. 

\begin{table*}[tb]
    \caption{\textbf{Study on generalization of temporal2Seq}. We compare the generalization performance from different pre-trained models on the FineAction, GTEA and TAPOS datasets. Temporal2Seq outperforms the other two for all datasets.}
    \label{table:generalization}
    \centering
    \resizebox{1.0\textwidth}{!}{
        \begin{tabular}{cccccccccccc}
            \hline
            \multirow{2}{*}{Model} & \multicolumn{4}{c}{TAD} & \multicolumn{5}{c}{TAS} & \multicolumn{2}{c}{GEBD}                                                                                                                                                            \\
            \cmidrule(r){2-5} \cmidrule(r){6-10}  \cmidrule(r){11-12}
                                   & 0.5                     & 0.75                    & 0.95                     & Avg            & \multicolumn{3}{c}{F1@\{10,25,50\}} & Edit           & Acc            & 0.05           & Avg                                            \\ \hline
            ViT-B                  & 25.54                   & 12.79                    & 3.45                     & 12.87          & 86.72                               & 83.47          & 70.23          & 83.42          & 76.28          & 63.9          & 67.8          \\
            Baseline$_{[TASK]}$    & 26.77                   & 13.52                    & 4.88                     & 13.78          & 89.44                               & 85.61          & 72.42          & 83.87          & 77.53          & 65.2          & 69.5          \\
            Temporal2Seq$_{batch}$  & \textbf{27.81}          & \textbf{13.92}           & \textbf{5.10}            & \textbf{14.12} & \textbf{89.82}                      & \textbf{88.76} & \textbf{74.15} & \textbf{84.63} & \textbf{78.08} & \textbf{65.5} & \textbf{70.0} \\ \hline
        \end{tabular}
    }
\end{table*}

\begin{table}[!htbp]
    \caption{\textbf{Comparison with recent task-specific models on three different tasks}. For a fair comparison, we use the ViT-B as the feature extractor for the current state-of-the-art non-end-to-end methods for each task. * indicates that we input randomly cropped video clips instead of the entire video to this model in order to align with Temporal2Seq. The statistics on model parameters do not include the parameters of the backbone network in the end-to-end method; only the parameters of the detectors in each method are calculated.
    }
    \label{table:sota}
    \centering
    \tabcolsep=0.01\textwidth
    \resizebox{1.0\textwidth}{!}{
        \begin{tabular}{llllcccccccccccc}
            \hline

            \multirow{2}{*}{Method}    & \multirow{2}{*}{Params(M)}                & \multirow{2}{*}{Backbone} & \multicolumn{6}{c}{TAD} & \multicolumn{5}{c}{TAS} & \multicolumn{2}{c}{GEBD}                                                                                                              \\
            \cmidrule(r){4-9} \cmidrule(r){10-14}  \cmidrule(r){15-16}
                               &                         &                                                 & 0.3                     & 0.4                     & 0.5                      & 0.6  & 0.7  & Avg-mAP & \multicolumn{3}{c}{F1@\{10,25,50\}} & Edit & Acc  & F1@0.05 & Avg-F1               \\ \hline
            \textbf{Specialist Models}     &             &                           &                                               &                         &                          &      &      &         &                                     &      &      &         &        &      &      \\
            RTD-Net~\citep{rtd}&  14.0                        & I3D                                             & 58.5                    & 53.1                    & 45.1                     & 36.4 & 25.0 & 43.6    & -                                   & -    & -    & -       & -      & -    & -    \\
            G-TAD~\citep{gtad}&6.1                           & TSN                                             & 66.4                    & 60.4                    & 51.6                     & 37.6 & 22.9 & 47.8    & -                                   & -    & -    & -       & -      & -    & -    \\
            TadTR~\citep{tadtr}&8.6                          & I3D                                             & 62.4                    & 57.4                    & 49.2                     & 37.8 & 26.3 & 46.6    & -                                   & -    & -    & -       & -      & -    & -    \\
            VSGN~\citep{vsgn}&8.4                            & TSN                                            & 66.7                    & 60.4                    & 52.4                     & 41.0 & 30.4 & 50.2    & -                                   & -    & -    & -       & -      & -    & -    \\
            AFSD~\citep{afsd}&14.2                            & I3D                              & 67.3                    & 62.4                    & 55.5                     & 43.7 & 31.1 & 52.0    & -                                   & -    & -    & -       & -      & -    & -    \\
            TALLFormer~\citep{tallformer}&15.0                & Swin-B                           & 76.0                    & -                       & 63.2                     & -    & 34.5 & 59.2    & -                                   & -    & -    & -       & -      & -    & -    \\
            ActionFormer~\citep{actionformer}&29.2            & I3D                                             & 82.1                    & 77.8                    & 71.0                     & 59.4 & 43.9 & 66.8    & -                                   & -    & -    & -       & -      & -    & -    \\
            ActionFormer~\citep{actionformer}&29.2            & ViT-B                                             &       79.7              &     74.3                &   68.2                  &  57.3& 42.3 & 64.3    & -                                   & -    & -    & -       & -      & -    & -    \\
            
            MS-TCN~\citep{mstcn}&0.8                         & I3D                                             & -                       & -                       & -                        & -    & -    & -       & 58.2                                & 52.9 & 40.8 & 61.4    & 65.1   & -    & -    \\
            MS-TCN++~\citep{mstcn++}&0.8                     & I3D                                             & -                       & -                       & -                        & -    & -    & -       & 64.1                                & 58.6 & 45.9 & 65.6    & 67.6   & -    & -    \\
            ASRF~\citep{asrf}&1.2                          & I3D                                             & -                       & -                       & -                        & -    & -    & -        & 74.3 & 68.9 & 56.1  & 72.4  & 67.6   & -    & -    \\
            UARL~\citep{uarl}&1.5                          & I3D                                             & -                       & -                       & -                        & -    & -    & -        & 65.2 & 59.4 & 47.4  & 66.2  & 67.8   & -    & -    \\
            HASR~\citep{hasr}&19.2                          & I3D                                             & -                       & -                       & -                        & -    & -    & -        & 74.7 & 69.5 & 57.0  & 71.9  & 69.4   & -    & -    \\
            UVAST~\citep{uvast}&1.3                          & I3D                                            & -                       & -                       & -                        & -    & -    & -       & 76.9                                & 71.5 & 58.0 & 77.1    & 69.7   & -    & -    \\
            ASFormer~\citep{asformer}&1.1                    & I3D                                             & -                       & -                       & -                        & -    & -    & -       & 76.0                                & 70.6 & 57.4 & 75.0    & 73.5   & -    & -    \\
            DiffAct~\citep{diffact}&1.2                     & I3D                                             & -                       & -                       & -                        & -    & -    & -       & 80.3                                & 75.9 & 64.6 & 78.4    & 76.4   & -    & -    \\
            DiffAct~\citep{diffact}&1.2                      & ViT-B                                             & -                       & -                       & -                        & -    & -    & -       & 78.7                                & 74.9 & 62.7 & 76.4    & 74.9   & -    & -    \\
            
            BMN-StartEnd~\citep{bmn}&3.1                     & ResNet50                                        & -                       & -                       & -                        & -    & -    & -       & -                                   & -    & -    & -       & -      & 49.1 & 64.0 \\
            TCN-TAPOS~\citep{tcn}&0.8                        & ResNet50                                        & -                       & -                       & -                        & -    & -    & -       & -                                   & -    & -    & -       & -      & 46.4 & 62.7 \\
            TCN~\citep{tcn}&0.8                              & ResNet50                                        & -                       & -                       & -                        & -    & -    & -       & -                                   & -    & -    & -       & -      & 58.8 & 68.5 \\
            Temporal Perceiver~\citep{temporalperceiver}&52.4 & ResNet50                                        & -                       & -                       & -                        & -    & -    & -       & -                                   & -    & -    & -       & -      & 74.8 & 86.0 \\
            Temporal Perceiver~\citep{temporalperceiver}&52.4 & ViT-B                                        & -                       & -                       & -                        & -    & -    & -       & -                                   & -    & -    & -       & -      & 75.2 & 85.7 \\
            
            BasicGEBD~\citep{efficientgebd}&1.0 & ResNet50                                           & -                       & -                       & -                        & -    & -    & -       & -                                   & -    & -    & -       & -      & 76.8 & 86.6 \\
            EfficientGEBD~\citep{efficientgebd}&1.0 & ResNet50                                           & -                       & -                       & -                        & -    & -    & -       & -                                   & -    & -    & -       & -      & 78.3 & 88.3 \\
            Baseline$_{TAD}$&6.9                            & ViT-B                                           & 80.6                    & 74.2                    & 66.3                     & 53.1 & 37.3 & 62.3    & -                                   & -    & -    & -       & -      & -    & -    \\
            Baseline$_{TAS}$&6.9                            & ViT-B                                           & -                       & -                       & -                        & -    & -    & -       & 75.7                                & 70.8 & 57.8 & 73.4    & 70.7   & -    & -    \\
            Baseline$_{GEBD}$&6.9                           & ViT-B                                           & -                       & -                       & -                        & -    & -    & -       & -                                   & -    & -    & -       & -      & 75.0 & 85.5 \\ \hline
            \textbf{Generalist Models}                                            &                      &                         &                         &                          &      &      &         &                                     &      &      &         &        &      &      \\
            Temporal2Seq$_{batch}$&6.9                      & ViT-B                                           & 79.9                    & 74.6                    & 67.4                     & 56.4 & 41.9 & 64.0    & 77.5                                & 71.6 & 59.7 & 75.0    & 72.2   & 75.2 & 86.4 \\
            Temporal2Seq$_{data}$&6.9                       & ViT-B                                           & 80.3                    & 74.2                    & 67.7                     & 55.2 & 40.3 & 63.5    & 77.0                                & 71.5 & 60.3 & 74.7    & 71.8   & 75.6 & 86.5 \\ \hline
        \end{tabular}
    }
\end{table}

\noindent\textbf{Study on the Generalization of Temporal2Seq.}
To further explore the effectiveness of a generalist model from joint training, we verify its transfer ability to a new dataset for each task. Here we treat the following three models as the pre-trained models that are fully fine-tuned on three unseen datasets (FineAction, GTEA and TAPOS) for each task: \textbf{ViT-B} pre-trained on Kinetics-400~\citep{k400}, \textbf{Baseline$_{[TASK]}$} trained on one dataset for each task (THUMOS14 for TAD, Breakfast for TAS and Kinetics-GEBD for GEBD), and \textbf{Temporal2Seq$_{data}$} trained on all three datasets.
Shown in Table~\ref{table:generalization}, with the introduction of each dataset belonging to each task, the model has learned the priors from each specific dataset. So the results of \textbf{Baseline$_{[TASK]}$} are better than \textbf{ViT-B}. 
Furthermore, the results are further improved on all three new datasets based on \textbf{Temporal2Seq$_{batch}$}, demonstrating the better generalization ability. 

\subsection{Comparison with the State-of-the-Art Methods}
We compare Temporal2Seq with other task-specific methods across three tasks shown in Table~\ref{table:sota}. Since some methods in TAD~\citep{actionformer} and all methods in TAS input the whole video into the model while Temporal2Seq requires random sampling to balance the joint training of different tasks. We set $L=1200$ to ensure almost all actions can be captured for all three tasks to make a fair comparison with other SOTA methods.
Shown in Table~\ref{table:sota}, our method outperforms a large proportion of existing specialist models. Since Temporal2Seq uses a simple Transformer-based modeling module and, for the sake of decoder unification, does not incorporate complex decoding structures like other specialist models, its performance falls short compared to the best specialist models. However, this gap is considered acceptable. Our Temporal2Seq achieves comparable performance with one of the best TAD methods called Actionformer when they both adopt ViT-B as their backbones. It also comparable with one of the best TAS methods called DiffAct and outperforms Temporal Perceiver when both adopt ViT-B as their backbones. Also, Temporal2Seq's flexible decoding structure, which is divided into dense and sparse types, can be adapted to a wider range of tasks and datasets. Moreover, its straightforward data balancing strategy requires no complex design—training can be accomplished simply by adjusting the batch proportions. In terms of parameter count, most specialist models use only lightweight detection heads, especially most methods in the TAS and GEBD tasks. However, the TAD methods have a significantly higher parameter count due to their more complex boundary regression approach. Temporal2Seq's parameter count falls between these methods and is acceptable.
Above all, Temporal2Seq addresses three fundamental temporal video understanding tasks with a single model. Its approach can be extended to more temporal video understanding tasks, leveraging diverse datasets to enhance the model's understanding capabilities.

\section{Conclusion and Limitation}
In this paper, we have proposed a single unified unimodal framework of Temporal2Seq for dealing with different video understanding tasks. Temporal2Seq formulates the output of each task as a sequence of discrete tokens, which enables a unified interface via a task prompt to three tasks without consideration of designing complex heads for each task. We successfully co-train our Temporal2Seq model on three diverse video understanding tasks, covering detection, segmentation, and timestamp localization, and the experiment results empirically demonstrate the improvement of co-training across all three tasks over each individual-specific model. We also show the promising generalization ability of our trained generalist model on new datasets among these tasks. 

The main limitation of Temporal2Seq lies in its inability to match the performance of task-specific specialist models, which stems from its relatively simple architecture. However, designing complex structures is not the focus of this work. The most significant contribution of Temporal2Seq is that it overcome the performance bottleneck in joint training for downstream fine-grained temporal tasks. Through its data balancing strategy, it prevents the joint training from being affected by scale disparities across different datasets.

Temporal2Seq has already demonstrated its ability to understand similar semantics from various temporal video understanding tasks, and we look forward to more future works to improve it.

\printcredits

\section*{Declaration of competing interest}
The authors declare that they have no known competing financial interests or personal relationships that could have
appeared to influence the work reported in this paper.

\section*{Data availability}
All the datasets used in this study are open-source academic datasets and are publicly available. The code and models will be open-sourced.

\section*{Acknowledgment}

This work is supported by the Frontier Technologies R\&D Program of Jiangsu (No. BF2025008) and the Collaborative Innovation Center of Novel Software Technology and Industrialization.

\bibliographystyle{cas-model2-names}




\end{document}